%% file: main.tex
\documentclass[final]{cvpr}

\usepackage{times}
\usepackage{epsfig}
\usepackage{graphicx}
\usepackage{amsmath}
\usepackage{amssymb}

\usepackage{colortbl}
\usepackage{comment}
\usepackage{bm}
\usepackage{myfigures}
\usepackage{mytables}
\usepackage{algorithm, algpseudocode}
\usepackage{mathtools}
\usepackage{arydshln}
\usepackage{tabularx}
\newcolumntype{C}{>{\centering\arraybackslash}X}

\usepackage{multirow}

\usepackage{capt-of}

\newcommand{\child}{}
\newcommand{\parent}{^{\prime}}
\newcommand{\dir}{\bm{\delta}}
\newcommand{\dirs}{\bm{\Delta}}

\usepackage[table,xcdraw,dvipsnames]{xcolor}

\usepackage[pagebackref=true,breaklinks=true,colorlinks,bookmarks=false]{hyperref}

\def\cvprPaperID{2178} 
\def\confYear{CVPR 2021}
\begin{document}

\title{Heterogeneous Grid Convolution for\\ Adaptive, Efficient, and Controllable Computation}

\author{Ryuhei Hamaguchi$^{1}$, Yasutaka Furukawa$^{2}$, Masaki Onishi$^{1}$, and Ken Sakurada$^{1}$ \\
$^{1}$ National Institute of Advanced Industrial Science and Technology (AIST)\\
$^{2}$ Simon Fraser University \\
{\tt\small ryuhei.hamaguchi@aist.go.jp, furukawa@sfu.ca, onishi@ni.aist.go.jp, k.sakurada@aist.go.jp}
}

 \twocolumn[{
 \maketitle
 \vspace{-2em}
 \centerline{
$\begin{array}{cccc}\scriptsize
  \hspace{0mm} \includegraphics[width=4.2cm]{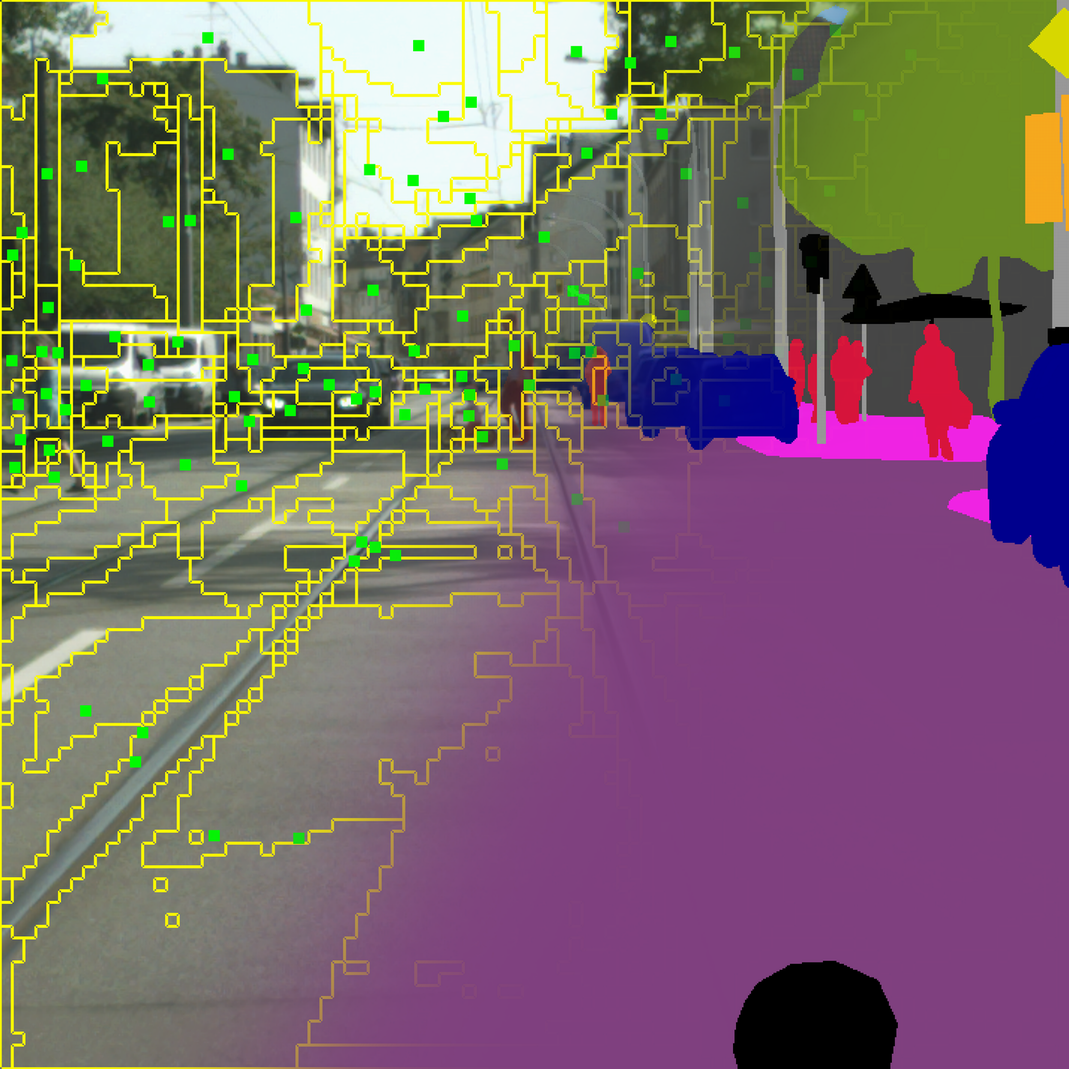} 
&
  \hspace{-2mm} \includegraphics[width=4.2cm]{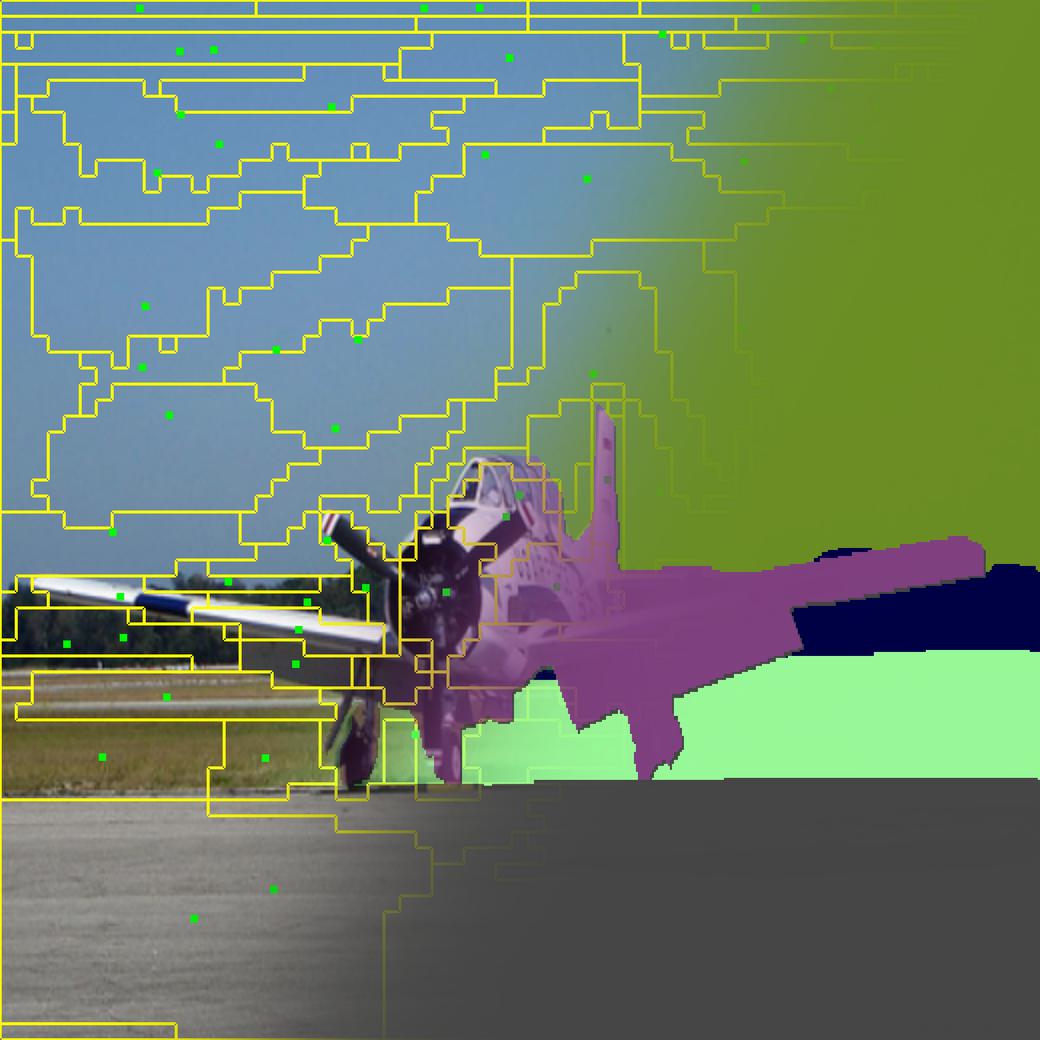} 
&
  \hspace{-2mm} \includegraphics[width=4.2cm]{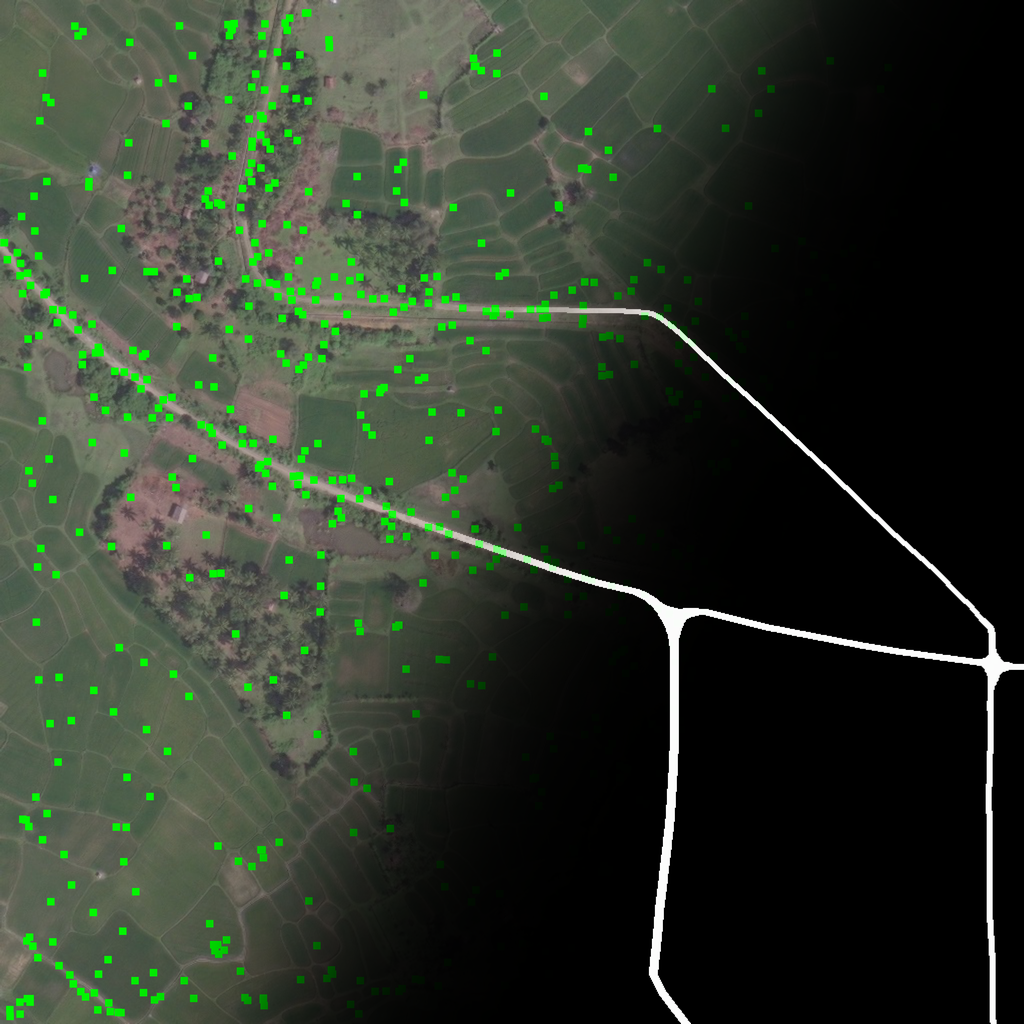}
&
  \hspace{-2mm} \includegraphics[width=4.2cm]{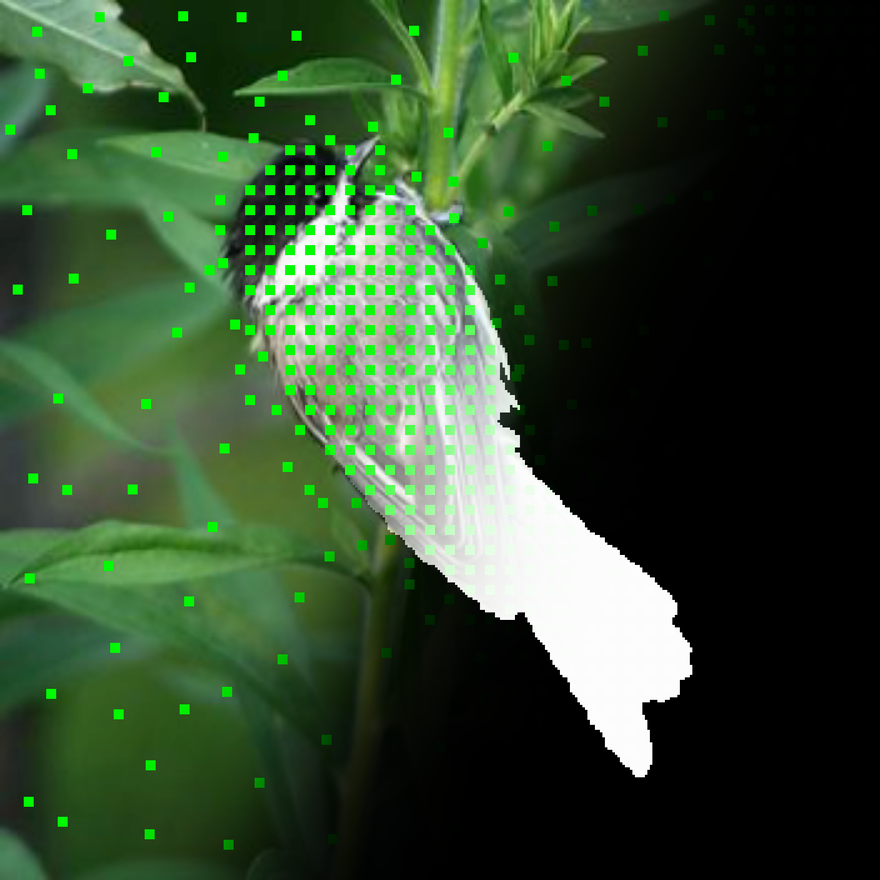}
\end{array}$
}
\captionof{figure}{
Heterogeneous grid convolution exploits the heterogeneity in the image to enable adaptive, efficient, and controllable computation for a range of image understanding tasks such as semantic segmentation, road extraction, and salient object detection from left to right.
}
\label{fig:teaser}
 \vspace{1em}
 }]

\input{0_abstract}

\input{1_introduction}

\figHGConv
\input{2_related_work}

\input{3.5_preliminaries}

\figAdj

\input{3_method}

\figNoiseCanceling

\figVisualization

\input{4_experiments}

\input{5_conclusion}

{\small
\bibliographystyle{ieee_fullname}
\bibliography{main}
}

\newpage
\appendix
\input{6_appendix}

\end{document}

%% file: 0_abstract.tex
\begin{abstract}
This paper proposes a novel heterogeneous grid convolution that builds a graph-based image representation by exploiting heterogeneity in the image content, enabling adaptive, efficient, and controllable computations in a convolutional architecture.
More concretely, the approach builds a data-adaptive graph structure from a convolutional layer by a differentiable clustering method, pools features to the graph, performs a novel direction-aware graph convolution, and unpool features back to the convolutional layer.
By using the developed module, the paper proposes heterogeneous grid convolutional networks, highly efficient yet strong extension of existing architectures.
We have evaluated the proposed approach on four image understanding tasks, semantic segmentation, object localization, road extraction, and salient object detection. The proposed method is effective on three of the four tasks. Especially, the method outperforms a strong baseline with more than 90\% reduction in floating-point operations for semantic segmentation, and achieves the state-of-the-art result for road extraction.
We will share our code, model, and data.
\end{abstract}

%% file: 1_introduction.tex
\section{Introduction}
\label{sect:introduction}

Our world is heterogeneous in nature. Looking at a scene from a car (See Fig.~\ref{fig:teaser}), the road occupies one third of the image with homogeneous textures. At the far end of the road are full of objects such as cars, pedestrians, or road signs. The density of semantic information varies per location. Our attention to the world is also heterogeneous in nature. With a specific task in mind, we focus our attention to a specific portion of an image, for example, tracing a road network in a satellite image.

While a regular grid feature representation has been successful, such a representation contains redundant information in low density regions, whereas the spatial resolution is insufficient in high density regions. Features should be stored adaptively based on the information density.

This paper studies a novel ``heterogeneous grid convolution'', which has the following three advantages.
({\it Adaptive}) The node features are adaptively allocated where necessary.
({\it Efficient}) The adaptive allocation reduces redundant computations.
({\it Controllable}) An agent can focus computation to a region of interest with an additional input.


There are two technical challenges in learning such a flexible feature representation:
(a) how to adaptively allocate nodes while exploiting image heterogeneity, and
(b) how to define a convolution operation on a heterogeneous grid structure.
We propose a differentiable clustering-based graph pooling for the first challenge and a direction-aware extension of the graph convolution for the second challenge.

The combination of our graph pooling and direction-aware graph convolution forms a neural model, dubbed heterogeneous grid convolution, that can be inserted into any existing CNN architecture.
By exploiting the proposed module, we also propose a heterogeneous grid convolutional neural networks (HG-CNNs) as a highly efficient yet strong extension of existing CNN architectures.

We have evaluated the proposed approach on four image understanding tasks, semantic segmentation, object localization, road extraction, and salient object detection.
The proposed HG-CNNs are effective for three of the four tasks; the HG-CNN outperforms strong baselines with fewer floating-point operations (more than 90\% reduction) on semantic segmentation; it achieves the state-of-the-art result on road extraction task; it yields compelling performance against state-of-the-arts on salient object detection.
The current neural processors (i.e., NPU and GPU) are optimized for regular grid computations, and the proposed module is not necessarily computationally faster or more efficient in practice.
However, the paper opens up a new avenue of research, potentially leading to an impact in vertical application domains, such as embedded devices with specialized hardware.
We will share all our code and data to promote further research.

%% file: 2_related_work.tex
\section{Related works}
\label{sect:related_works}

The literature of convolutional neural architecture is massive. The section focuses the description on the graph convolution, the graph pooling, and other closely related enhancement techniques in computer vision.

\vspace{0.1cm}
\noindent
\textbf{Graph convolution:}
Hammond et al. \cite{Hammond2011} and Defferrard et al. \cite{Defferrard2016} formulated a convolution operation on graph-structured data based on spectral graph theory, approximating filters in the Fourier domain using Chebyshev polynomials. Kipf and Welling \cite{Kipf2017} further proposed a first-order approximation of the spectral graph convolution. Since the above works assume general graph data as inputs, they lack the capability of capturing spatial relationships between nodes for embedded graphs.
To remedy this, various methods have been proposed \cite{Fey2018,Li2019,Monti2017,Spurek2020,Wang2019,You2019}. For instance, Spline-CNN \cite{Fey2018} extends a regular kernel function to a continuous kernel function using B-spline bases, where convolution weights for the adjacent nodes can be modified according to their relative spatial position. In our experiments, we compare our direction-aware graph convolution to the above methods.

Ci et al. ~\cite{Ci2019} extends widely used GCN for 3D pose estimation by using different weight parameters for every pair of nodes. However the application of the method is limited to the tasks where the graph structure is pre-defined, e.g., a skeleton body model in 3D pose estimation.

\vspace{0.1cm}
\noindent
\textbf{Graph pooling:}
Graph pooling is a key operation for learning hierarchical graph representations. DiffPool was proposed as a differentiable graph pooling method, in which soft-cluster assignments are directly estimated using graph convolution layers in an end-to-end manner~\cite{Ying2018}.
Other methods defined graph pooling as a node selection problem. In such methods, the top-k representative nodes are selected using a trainable vector $p$ \cite{Cangea2018} or self-attention mechanism \cite{Lee2019}. Whereas the above methods globally select discriminative nodes, AttPool \cite{Huang2019} also applies a local attention mechanism to prevent the pooling operation from being stuck within a narrow sub-graph.

\vspace{0.1cm}
\noindent
\textbf{Non-grid representations in computer vision:}
Graph-based representations have been proposed for modeling long-range dependencies in an image. Li et al. proposed a module that performs graph reasoning on a fully-connected graph acquired from a clustering-based graph projection~\cite{Li2018}.
Similar ideas are also proposed by \cite{Chen2019} and \cite{Zhang2019}. To reduce the computational complexity of the fully-connected graph reasoning, recent work proposed a dynamic graph message passing that adaptively constructs a local graph for each location of a feature map \cite{Zhang2020}.
These methods aim to refine a regular grid representation by adding an extra graph reasoning module on it, and thus still depend on regular convolution for spatial feature extraction.
On the other hand, our aim is to replace the redundant regular convolutions by the proposed HG-Conv that gives a unified method for spatial feature extraction and long-range message passing on compact graph representations.

Marin et al. proposed a non-uniform downsampling method that learns deformations from uniform sampling points such that the points near semantic boundaries are sampled as many as possible \cite{Marin2019}.
More recently, Gao et al. proposed a method that constructs an adaptive triangular mesh on an image plane, and applied the method as learnable downsampling on semantic segmentation task \cite{Gao2020}. The method predicts deformations from an initial triangular mesh such that each mesh has a small feature variance.
These methods differ from our method in two points; 1) they applied conventional regular convolutions after non-uniform downsampling; 2) for this reason, the deformations are restricted so that the regularity of the output are kept. For this purpose, the methods introduced regularization terms.
On the other hand, our graph convolution can operate directly on non-uniform inputs, and hence the proposed graph pooling can generate pooling regions of arbitrary shapes and sizes in a purely data adaptive manner.

PointRend \cite{Kirillov2020} is proposed as a point-based feature refinement module. To generate high-resolution output, the module adaptively samples points from upsampled feature maps and apply MLP to refine the point features. The method is orthogonal to our method.

Ning et al. ~\cite{Ning2019} proposed an efficient convolution method by reusing computation among similar features. While the method achieves an efficient approximation of a regular convolution, the method cannot be applied on non-grid inputs.

\vspace{0.1cm}
\noindent
\textbf{Other enhancement techniques:}
Dilated convolutions \cite{Yu2016,Yu2017} take a strategy of maintaining the feature resolution throughout the networks.
Despite being a successful technique ~\cite{Yuan2019,Yuan2018,Zhang2019ACFNet,Zhao2017}, Dilated convolutions suffer from large memory consumption and computations due to the high-resolution feature maps.
More recently, multi-resolution features~\cite{Wang2020} or a course-to-fine strategy~\cite{Badrinarayanan2015,Chen2018deeplab,Lin2018,Lin2017,Noh2015,Ronneberger2015} have been proposed to alleviate the issue.

Multi-scale feature fusion has been studied for aggregating long-range contextual information~\cite{Chen2018deeplab,He2019,Yang2018,Yu2016,Zhao2017}.
The methods build multi-scale features by applying pyramid pooling \cite{Zhao2017} or dilated convolutions with different dilation rates~\cite{Chen2018deeplab}.
Recent works \cite{Chen2018A2Net,Yuan2019,Yuan2018,Zhang2019ACFNet,Zhang2019CFNet} have proposed adaptive context aggregation methods that are based on the feature relation. For instance,  OCNet \cite{Yuan2018} identifies the context for each pixel by adopting a self-attention mechanism. $A^2$-Net \cite{Chen2018A2Net} applies a double attention mechanism, where the key features in a scene are aggregated during the first ``gather'' attention, and are distributed to each pixel during the second ``distribute'' attention.

%% file: 3.5_preliminaries.tex
\section{Convolution as a set of graph-convolutions}
\label{sect:preliminaries}

Convolution is a direction-wise set of graph-convolutions. We first show this not well-known fact, which will allow us to define heterogeneous grid convolution with the language of graph-convolutions towards a simple and efficient implementation in the next section.

Considering convolution as a message-passing architecture, (3$\times$3) convolution passes messages along nine directions $\dirs=\left\{\leftarrow,\rightarrow,\uparrow,\downarrow,\nwarrow,\nearrow,\swarrow,\searrow,\circlearrowleft\right\}$
(See Fig.~\ref{fig:da_graph_conv}):
\begin{equation}
    \overrightarrow{\bm{z}_{\bm{p}}} = \sum_{\dir\in \dirs}\overrightarrow{\bm{x}_{\bm{p}+\dir}}\bm{W}_{\dir}.
\label{eq:regular_conv}
\end{equation}
$\overrightarrow{\bm{x}_{\bm{p}}}$ is the (1$\times \mathcal{N}_{in}$) input feature vector at pixel $\bm{p}$. $\overrightarrow{\bm{z}_{\bm{p}}}$ is the (1$\times \mathcal{N}_{out}$) output feature vector. With abuse of notation $\dir(\in \dirs)$ is a positional displacement for a given direction.
$\bm{W}_{\dir}$ is the ($\mathcal{N}_{in}\times \mathcal{N}_{out}$) kernel matrix for direction $\dir$.~\footnote{A kernel set is a 4D tensor, usually interpreted as a 2D matrix for a pair of input and output channels. 
$\bm{W}_{\dir}$ is a 2D slice of the 4D tensor per pixel, while masking out the contributions outside the given direction $\dir$.}

Let $\bm{X}$ and $\bm{Z}$ denote the set of feature vectors for all the pixels as the ($\mathcal{N}_{pix}\times \mathcal{N}_{in}$) and ($\mathcal{N}_{pix}\times \mathcal{N}_{out}$) matrices, where $\mathcal{N}_{pix}$ is the number of pixels. The above message-passing equation can be written for all the pixels as
\begin{equation}
    \bm{Z} = \sum_{\dir\in \dirs}\left(\bm{D}^{\dir}\right)^{-1}
    \bm{A}^{\dir}
    \bm{X}\bm{W}_{\dir}.
\label{eq:direction-aware_graph_conv}
\end{equation}
$\bm{A}^{\dir}$ is the ($\mathcal{N}_{pix} \times \mathcal{N}_{pix}$) asymmetric adjacency matrix for direction $\dir$, that is, $\bm{A}_{ij}^{\dir}$ is 1 if the $i_{\mbox{th}}$ pixel is connected to the $j_{\mbox{th}}$ pixel along direction $\dir$.
$\bm{D}^{\dir}$ is the ($\mathcal{N}_{pix} \times \mathcal{N}_{pix}$) degree matrix of $\bm{A}^{\dir}$: $\bm{D}^{\dir}_{ii} = \max\left(\sum_j \bm{A}^{\dir}_{ij},\epsilon\right)$, where ($\epsilon=$1e-7) is used to avoid divide-by-zero in computing its inverse.
The formula inside the summation is the widely used graph-convolution formulation by Kipf and Welling \cite{Kipf2017}, which is summed over the message passing directions.

%% file: 3_method.tex
\section{Heterogeneous Grid Convolution}
\label{sect:method}

Heterogeneous grid convolution (HG-Conv) is a natural extension of convolution in the heterogeneous grid domain. Understanding that the convolution is equivalent to the sum of direction-wise graph-convolutions (Eq.~\ref{eq:direction-aware_graph_conv}), HG-Conv is defined as a four-step process shown in Fig.~\ref{fig:hgconv}:
(1. Clustering) Find groups of pixels sharing similar features; (2. Pooling) Compute the group feature vector by taking the average over its pixels; (3. Graph-convolution) Perform convolution as direction-wise graph-convolutions over the groups; and (4. Unpooling) Copy the group feature vector back to the pixels for each group.
The four steps are defined in the following formula: 
\begin{eqnarray}
    \bm{Z} &=& \bm{S}\sum_{\dir \in \dirs} \left(\bm{\hat{D}}^{\dir}\right)^{-1} \bm{\hat{A}}^{\dir} \bm{S}^T \bm{X}\bm{W}_{\dir},\\
    \bm{\hat{A}}^{\dir} &=& \bm{S}^T\bm{A}^{\dir}\bm{S}.
\label{eq:heterogeneous_grid_conv}
\end{eqnarray}
$\bm{S}$ is a $\mathcal{N}_{pix}\times \mathcal{N}_{grp}$ group assignment matrix, where $\bm{S}_{pg}$ defines an assignment weight from pixel $p$ to group $g$.
$\bm{\hat{A}}^{\dir}$ is the $\mathcal{N}_{grp}\times \mathcal{N}_{grp}$ adjacency matrix for the groups.
$\bm{\hat{D}}^{\dir}$ is the $\mathcal{N}_{grp}\times \mathcal{N}_{grp}$ degree matrix of $\bm{\hat{A}}^{\dir}$.
Starting from the stack of input feature vectors $\bm{X}$, (1. Clustering) is to compute $\bm{S}$; (2. Pooling) is the left-multiplication of $\bm{S}^T$; (3. Graph-convoluion) is the left-multiplication of  $(\bm{\hat{D}}^{\dir})^{-1} \bm{\hat{A}}^{\dir}$ and right-multiplication of the learnable kernel $\bm{W}_{\dir}$; and (4. Unpooling) is the multiplication of $\bm{S}$.

\subsection{Differentiable clustering}
The group assignment $\bm{S}$ is computed by sampling cluster centers from input pixels, and associating input features to the cluster centers using differentiable SLIC algorithm~\cite{Jampani2018}. Note that $\bm{S}$ is a soft-assignment and trainable in an end-to-end manner.
The cluster centers are sampled based on ``importance'' of each pixel. The importance is defined as $L^2$ distance between a pixel's feature and its adjacent features.
As an extension, the importance map can be incorporated as an attention map for controlling node allocation as shown later.

\subsection{Pooling}
Given the group of pixels, group feature vectors are computed by the average pooling, which can be written as:
\begin{equation}
    \bm{\hat{X}}=\bm{\bar{S}}^{T}\bm{X}\child.
\label{eq:update_x}
\end{equation}
$\bm{\bar{S}}$ is a column-wise normalized assignment matrix, i.e., $\bm{\bar{S}}=\bm{S}\bm{\bar{Z}}^{-1}$ and $\bar{Z}_{jj}=\sum_{i}S_{ij}$.
The unpooling operation is defined via its transpose:
\begin{equation}
    \bm{X}\child=\bm{\tilde{S}}\bm{X}\parent
\label{eq:unpool_x}
\end{equation}
where $\bm{\tilde{S}}$ is a row-wise normalized matrix, i.e., $\bm{\tilde{S}}=\bm{\tilde{Z}}^{-1}\bm{S}$ and $\tilde{Z}_{ii}=\sum_{j}S_{ij}$.

\subsection{Graph-convolution}
A convolution (Eq.~\ref{eq:direction-aware_graph_conv}) is defined as the left multiplication of the adjacency matrix $\bm{A}^{\dir}$ (with the inverse of the degree matrix) and the right multiplication of the learnable kernel $\bm{W}_{\dir}$. 
We define a convolution for groups by simply replacing the adjacency matrix of the pixels $\bm{A}^{\dir}$ with the adjacency matrix of the groups $\bm{\hat{A}}^{\dir}$.
$\bm{\hat{A}}^{\dir}_{ij}$ should encode the amount of connections from the $i_{th}$ group to the $j_{th}$ group along direction $\dir$, in other words, how many pixel-level connections there are along $\dir$ from a pixel in the $i_{th}$ group to a pixel in the $j_{th}$ group. This can be calculated easily by the group assignment matrix S: $\bm{\hat{A}}^{\dir} = \bm{S}^T\bm{A}^{\dir}\bm{S}$~\cite{Ying2018}.
The convolution for the groups is then given as the left multiplication of $(\bm{\hat{D}}^{\dir})^{-1} \bm{\hat{A}}^{\dir}$ and the right-multiplication of $\bm{W}_{\dir}$.

We find that the clusters from differentiable SLIC tend to have complicated shapes and include many small disjoint regions, where
Fig.~\ref{fig:graph_diffslic}~(a) illustrates the situation. Due to the disjoint cluster (depicted in blue), the green cluster is connected to the blue cluster in every direction, which results in a ``noisy'' group adjacency matrix (Fig.~\ref{fig:graph_diffslic}~(b)).

To this end, a ``noise-canceling operation'' is performed on the group adjacency matrix, which cancels out the connection weight by the weight of the opposite direction (Fig.~\ref{fig:graph_diffslic}~(c)). Let $\bar{\dir}$ be the opposite direction of $\dir$, the ``noise-canceling'' is performed as follows.
\begin{equation}
{\bm{\hat{A}}}^{\dir}\leftarrow{\rm max}(\bm{0}, {\bm{\hat{A}}}^{\dir}-{\bm{\hat{A}}}^{\bar{\dir}})
\end{equation}
By abuse of notation, ${\bm{\hat{A}}}^{\dir}$ is replaced with the refined matrices. In practice, the matrices are further simplified by only keeping the direction with the maximum connection (e.g., only the left-wise connection remains in case of Fig.~\ref{fig:graph_diffslic}~(c)). We empirically find that taking the strongest direction slightly improve the performance.

\subsection{HG convolutional modules and networks}
A standard convolutional architecture design is to repeat convolutional, batch normalization, and ReLU layers. This design is immediately applicable to HG convolutions:
\begin{eqnarray}
    \bm{Z}_{0}&=&\bm{S}^{T} \bm{X},
    \label{eq:hgmodule_pool} \\
    \bm{Z}_{l}&=&F_{\rm BN\mathchar`-ReLU}\left(
    \sum_{\dir \in \dirs} \left(\bm{\hat{D}}^{\dir}\right)^{-1} \bm{\hat{A}}^{\dir}
    \bm{Z}_{l-1}
    \bm{W}_{\dir}
    \right),
    \label{eq:hgmodule_conv} \\
    \bm{Z}&=&\bm{S}\bm{Z}_{L}.
\label{eq:hgmodule_unpool}
\end{eqnarray}
This HG-Conv module repeats HG-convolutions, batch normalization, and ReLU L times from $\bm{X}$ to $\bm{Z}$.
$F_{\rm BN\mathchar`-ReLU}$ is a batch normalization layer followed by ReLU activation.
HG convolution is capable of incorporating other popular modules in the CNN literature such as residual blocks.
Next, we will design a few representative heterogeneous convolutional neural networks (HG-CNNs) by using HG-convolution and other techniques, where the full specifications are referred to the supplementary.

\vspace{0.1cm}
\noindent
\textbf{HG-ResNet:}
A ResNet \cite{He2016} is extended by replacing the 4th stage of the network with the HG-Conv module: The pooling (Eq.~\ref{eq:hgmodule_pool}) is inserted at the beginning of stage 4; The subsequent regular convolutions are replaced by the HG-Conv; and The unpooling (Eq.~\ref{eq:hgmodule_unpool}) is inserted at the end of the stage. Finally, the module output is concatenated to the stage 3 output and further refined by a $1\times 1$ convolution. Note that the parameter size is equal to the original ResNet except the final $1\times 1$ convolution.

\vspace{0.1cm}
\noindent
\textbf{HG-HRNetV2:}
HRNetV2 \cite{Wang2020} is a variant of HRNet that have recently shown outstanding performance on semantic segmentation. Similar to HG-ResNet, the HG-Conv module is applied to the last part of HRNetV2, in particular, all the 4 branches at the last block of stage 4.

\vspace{0.1cm}
\noindent
\textbf{HG-ResUNet:}
ResUNet is a variant of UNet \cite{Ronneberger2015}, popular in the field of medical image analysis and remote sensing. We applied the HG-Conv module to the last and the first block of the encoder and decoder, respectively. In the same way as HG-ResNet, the output of each module is concatenated with the input, and refined by $1\times 1$ convolution.


%% file: 4_experiments.tex
\section{Experiments}
\label{sect:experiments}
We evaluate the proposed HG-CNNs on four image understanding tasks, semantic segmentation, object localization, road extraction, and salient object detection.
On semantic segmentation, the HG-Conv outperforms strong baselines while representing an image with much fewer spatial nodes (less than 2\%) (Sect.~\ref{sect:exp_seg}).
However, HG-Conv does not perform effectively on object localization, which needs further exploration (Sect.~\ref{sect:exp_objloc}).
On the other two tasks, we demonstrate that the HG-Conv is able to control node allocations based on task-specific attention maps, an extension called ``active focus'' (Sects.~\ref{sect:exp_road} and \ref{sect:exp_sal}).

\subsection{Semantic Segmentation}
\label{sect:exp_seg}
\noindent
\textbf{Setup:}
We build three HG-CNNs based on HG-ResNet and HG-HRNetV2, and compare against their non-HG counterparts.
First, we use HG-ResNet to build two HG-CNNs (HG-ResNet-Dilation and HG-ResNet-DCN) by using dilated convolutions and deformable convolutions at the 3rd residual stage.
The non-HG counterparts (ResNet-Dilation and ResNet-DCN) are constructed by simply replacing the HG-Conv by dilated convolution and deformable convolution.
The third HG-CNN is HG-HRNetV2, where the non-HG counterpart is HRNetV2, which is the start-of-the-art segmentation network.
To further boost performance, we add auxiliary segmentation heads to the input of the HG-Conv modules for all HG-CNNs.

Unless otherwise noted, we determined the number of groups of HG-Conv as 1/64 of the number of input pixels (i.e., the downsampling rate is set as $1/64$).
As the HG-Conv adaptively constructs graph representations, the number of floating-point operations varies per image. Although the fluctuation is negligible, we evaluate the FLOPs of the HG-Conv by the average over the validation images. We basically used multi-scale prediction.

\vspace{0.1cm}
\noindent
\textbf{Main results:}
Fig.~\ref{fig:exp_seg} compares the performance and the computational complexity of the ResNet models and the HG-ResNet models with various depths (18/34/50/101).
The HG-ResNet models outperform the corresponding baselines with much less floating-point operations. Especially on PASCAL-context, HG-ResNet34-DCN outperforms ResNet101-DCN (+0.7\%) with only 10\% floating-point operations.
Furthermore, Table~\ref{tb:hrnet} shows that HG-HRNetV2 outperforms baseline HRNetV2 with less floating-point operations.

\vspace{0.1cm}
\noindent
\textbf{Comparison with other state-of-the-art non-grid convolutions}
Table~\ref{tb:comparison} shows the comparison against other state-of-the-art non-grid convolutions, DCN \cite{Dai2017} and DGMN \cite{Zhang2020}.
In the table, HG-Conv outperforms DCN with less FLOPs. The combination of DCN and HG-conv (DCN at stage 3 and HG-Conv at stage 4) outperforms DGMN with less FLOPs.
As for realistic runtime, HG-CNN is not faster than the conventional CNNs in practice. For instance, ``DCN'' in Table~\ref{tb:comparison} can process $713\times713$ inputs in 14.9 FPS, while ``DCN+HG-conv'' processes the same inputs in 6.5 FPS.  This is because today's processors (GPUs) are optimized for regular grid computations, not for graph processing. We believe that the runtime should be improved by more optimized implementation or specialized hardware.

In Table~\ref{tb:graphconv}, we also compared against other irregular convolutions studied in the field of geometric deep learning (e.g., GMMConv~\cite{Monti2017} and SplineConv~\cite{Fey2018}).
Specifically, we replace the graph-convolution step of the HG-Conv module with the competing modules (See Table~\ref{tb:graphconv}). The HG-Conv outperforms the other methods in most cases.
Due to engineering challenges, fine-tuning from ImageNet pre-trained models was not possible for some methods. For a fair comparison, we also trained our model from scratch.

\vspace{0.1cm}
\noindent
\textbf{Ablation study:}
To validate the design choices of the HG-Conv, we conducted several ablation studies.
\newline({\it Sampling methods}) 
We compare three sampling methods for the cluster center sampling step of the HG-Conv: random sampling, importance sampling, and a combination of top-k and random sampling \cite{Kirillov2020}.
Fig.~\ref{fig:visualization} visualizes the sampled cluster centers, and Table~\ref{tb:ablation_sampling} reports the model performances for each sampling method. With random sampling, a large portion of the sampled locations lie on the homogeneous road region, and many objects at the far end are missed.
In contrast, the other sampling methods properly place the cluster centers based on the importance map, which results in better segmentation performance.
\newline({\it Downsampling ratio}) Table~\ref{tb:ablation_downsampling} evaluates HG-ResNet with varying downsampling ratio. HG-ResNet outperforms the baseline ResNet with extremely small downsampling ratio.
\newline({\it Noise-canceling}) Table~\ref{tb:ablation_noise} demonstrates the effectiveness of noise-canceling operation on the adjacency matrices, which shows clear improvements on two of the three datasets. Max-direction heuristic achieves the well-balanced performance across all of the datasets. The effect of noise-canceling is qualitatively clear in Fig.~\ref{fig:visualization}.
\newline({\it HG-Conv for 3rd stage}) In Table~\ref{tb:ablation_stage}, we further convert the 3rd stage of ResNet101 into the HG-Conv. Whereas the performance degrades from 79.9\% to 78.1\%, the computational cost reduction
increases significantly (i.e., from 15.1 \% to 54.7 \%). However, as the result of HG-ResNet34 shows, reducing the depth of HG-ResNet is more effective than applying the HG-Conv at a shallow stage.

\figExpSeg

\subsection{Object Localization}
\label{sect:exp_objloc}
We also evaluated the proposed method on object localization tasks such as object detection and instance segmentation. For base models, we use Faster R-CNN and Mask R-CNN with FPN. Specifically, we compared two backbones, ResNet-DCN and HG-ResNet-DCN, on COCO dataset.

Table~\ref{tb:objloc} shows the results of the comparison. While our method achieves 30\% reduction in FLOPs, the accuracy of the model is degraded by HG-conv. We leave further exploration of HG-Conv on object localization for future work.

\subsection{Road Extraction}
\label{sect:exp_road}
Road extraction is a task for extracting road graphs from overhead images. On this task, we demonstrate the ``active focus'' capability that focuses cluster center allocation around predicted road lines.
Active focus is particularly effective for road extraction where targets have thin structure.
Finally, we apply the HG-Conv into the previous method, and achieve the state-of-the-art performance.

\tbHRNet
\tbComparison
\tbComparisonGraphConv
\tbAblationSampling
\tbAblationDownsampling
\tbSal
\tbAblationNoiseCanceling
\tbAblationStage

\tbObjLoc

\tbRoad

\vspace{0.1cm}
\noindent
\textbf{Setup:}
We make two HG-CNNs and their non-HG counterparts in this evaluation.
The first pair is HG-ResUNet and ResUNet, and the second pair is HG-Orientation and Orientation~\cite{Batra2019} (a current state-of-the-art network for the task).
We modified the above models slightly to keep high-resolution information; the stride of the first $7\times 7$ convolution is decreased to 1, and the max-pooling layer is removed ($^{+}$).
Furthermore, we make two modifications to employ active focus (-Attn): 1) A coarse segmentation head is added on the input feature of the HG-Conv module; and 2) The active focus is employed using the coarse prediction map as attention to focus the cluster center allocation on the road lines.

\vspace{0.1cm}
\noindent
\textbf{Results:}
Table~\ref{tb:road} shows that our method (HG-Orientation$^{+}$-Attn) achieves the state-of-the-art result on both IoU and APLS metrics. The effectiveness of HG-ResUNet18$^{+}$ is also clear (+0.6\% and +0.9\% for IoU and APLS, compared to ResUNet18$^{+}$).
The active focus is particularly effective on the task: By focusing spatial nodes on the road lines, the network can utilize high resolution information around road, while propagating contextual information from other regions (see Fig.~\ref{fig:visualization} for visualization).

\subsection{Salient Object Detection}
\label{sect:exp_sal}
Salient object detection is a task for identifying the object regions that are most attractive for human eyes. On the task, we compare two different types of active focus.

\vspace{0.1cm}
\noindent
\textbf{Setup:}
ResUNet50 and HG-ResUNet50 are used for the evaluations.
For active focus, the coarse segmentation head is attached on the HG-ResUNet50. We experiment two types of attention: ``object-aware'' and ``uncertainty-aware''. In object-aware attention, the predicted object mask is used as an attention, which places cluster centers on the object. In uncertainty-aware attention, entropy of the coarse prediction is used as an attention, which places cluster centers where the prediction is uncertain. For detailed training settings, please refer to the supplementary materials.

\vspace{0.1cm}
\noindent
\textbf{Results:}
Table~\ref{tb:sal} compares the HG-ResUNet models with baseline ResUNet50. For most datasets, HG-ResUNet50 outperforms the baseline. The object-aware active focus is particularly effective for the task. Whereas the uncertainty-aware active focus also performs well, the performance is worse compared to plain HG-ResUNet50 on three out of six datasets (See Fig.~\ref{fig:visualization}).

\vspace{0.1cm}
\noindent
\textbf{Comparison to the state-of-the-art methods:}
Fig~\ref{fig:exp_sal} compares the baseline and the proposed methods against the previous state-of-the-art. Our method does not achieve the state-of-the-art over all, but performs comparably well for some datasets without using complicated architectures such as iterative refinement modules \cite{Pang2020sod} or edge-aware loss functions \cite{Wu2019sod}.

\figExpSal

%% file: 5_conclusion.tex
\section{Conclusions}
This paper presents a novel heterogeneous grid convolution (HG-Conv) that builds an adaptive, efficient and controllable representation by exploiting heterogeneity inherent in natural scenes.
Our experimental results demonstrate that HG-CNN is capable of reducing computational expenses significantly without much sacrifice to performance, even achieving state-of-the-art for some tasks. HG-CNN is further capable of controlling the focus of computations based on an application-specific attention maps.
Our future work is to further explore the potentials of HG-CNN on more applications as well as the use of FPGA hardware, which is flexible and known to be effective for sparse computation.

\section*{Acknowledgement}
This paper is partially based on results obtained from a project commissioned by the New Energy and Industrial Technology Development Organization (NEDO) and supported by JSPS KAKENHI Grant Number 20H04217.

%% file: 6_appendix.tex
\section{Full architectural specifications}
\label{sect:settings}
Figure~\ref{fig:arch} provides the details of the four representative HG-CNN networks. For certain variants (specifically, HG-ResUNet18$^{+}$-Attn, HG-Orientation$^{+}$-Attn, and HG-ResUNet50-Attn), there exist some minor notes on the architecture and the following explains all the points.

\vspace{0.1cm}
\noindent\textbf{HG-ResUNet18$^{+}$-Attn (road extraction):}
1) The feature map from 4th residual stage of the encoder is skipped and concatenated with the output of HG-Conv modules on both of the encoder and decoder. (The concatenated features are further processed by $1\times 1$ convolution layers.)
2) We reuse the same assignment matrix $\bm{S}$ at the decoder (i.e., the clustering is only performed on the encoder.).

\vspace{0.1cm}
\noindent\textbf{HG-Orientation$^{+}$-Attn (road extraction):}
The major modification from the original model \cite{Batra2019} is 1) hourglass module is replaced by multiple branches of small hourglass modules, 2) HG-Conv is applied on the encoder and decoders in each branch, and 3) active focus is introduced on the second stack of the refinement part.

\vspace{0.1cm}
\noindent\textbf{HG-ResUNet50-Attn (salient object detection):}
The basic architecture is almost the same as HG-ResUNet18$^{+}$-Attn except several tiny differences as below; 1) The entry part is the same as the original ResNet (i.e., the stride of the first $7\times 7$ convolution is 2, and the max-pooling layer follows the convolution), 2) the dilated convolution is used at the 3rd and 4th residual stages. 3) the auxiliary loss is eliminated when the active focus is not used (i.e., HG-ResUNet50).

\section{Details on the noise canceling operation}
\label{sect:postproc}
The noise canceling operation has a few more post-processing steps on the group adjacency matrices $\bm{\hat{A}}^{\dir}$. Firstly, small connection weights ($\bm{\hat{A}}^{\dir}_{ij} < 10^{-7}$) are filtered out from the matrices (i.e., set to zero).
Secondly, self-loop $\hat{\bm{A}}^{\circlearrowleft}$ is always reset to an identity matrix.
Finally, the diagonal elements of the matrices $\hat{\bm{A}}^{\dir}$ are set to zeros except for the self-loop adjacency matrix $\hat{\bm{A}}^{\circlearrowleft}$.

\section{Details on the differentiable clustering}
\label{sect:clustering}
We use the differentiable SLIC algorithm~\cite{Jampani2018} in our architecture with one enhancement. SLIC samples initial cluster centers by uniform sampling. We use an adaptive algorithm from a prior work~\cite{Kirillov2020} for further improvements. The complete process is given in Algorithm~\ref{alg:diffslic}. The hyper-parameters of the algorithms are given in Table~\ref{tb:hyperparams}.

\algDiffSLIC

\section{Details on the importance map modulation}
\label{sect:focus}
Active focus modulates the importance map based on an application-specific attention map via simple weighted averaging. The section provides the full formula. 

Let $\bm{C}_{imp}$ denotes the ($H\times W$) importance map and $\bm{C}_{attn}$ denotes the ($H\times W$) attention map with its elements $(\bm{C}_{attn})_{ij}$ ranges from 0 to 1. The importance map is updated as follows:
\begin{equation}
    \bm{C}_{focus}=\frac{1}{Z}\bm{C}_{imp} + \alpha\bm{C}_{attn}.
\end{equation}
$Z=\max\left\{\bm{C}_{imp}[i,j]\right\}$ is a normalization coefficient.  
$\alpha$(=10) is a weight coefficient for the attention map.

As introduced in the main paper, we define two types of attention maps: object-aware and uncertainty-aware.
Let $\bm{P} = \{P_i | i = 1 \cdots K \}$ denote the $K$ class prediction map of shape ($K\times H\times W$). Object-aware active focus sets the attention map to $P_k$, where we are interested in an object in the $k_{th}$ class.
\begin{equation}
    \bm{C}_{attn} = \bm{P}_{k}.
\end{equation}
This form of attention focuses the cluster centers around the target class $k$.

Uncertainty-aware active focus computes the attention map as the entropy of the probability map.
\begin{equation}
    \bm{C}_{attn} = \frac{1}{\log{K}}\sum_{k}{\left\{-\bm{P}_{k}\log{\bm{P}_{k}}\right\}}.
\end{equation}
The above attention focuses the cluster centers on the regions where the model is uncertain for the prediction.

\tbDetailedResult

\section{Datasets and metrics}
\label{sect:examples}
\noindent\textbf{Semantic segmentation:}
To evaluate the proposed method, we use three semantic segmentation datasets, Cityscapes \cite{Cordts2016cityscapes}, ADE20K \cite{Zhou2017ade20k}, and PASCAL-context \cite{Mottaghi2014pascal}. For all the experiments, we use validation sets to evaluate the models.

\noindent $\bullet$
\textbf{Cityscapes} is a dataset for urban scene parsing, which contains 5,000 images of resolution $1,024\times 2,048$ with fine pixel-wise annotations. The annotations have 30 classes. We use the major 19 classes by following a prior convention. The dataset has the training, validation, and testing sets with 2,975/500/1,525 images, respectively. Only fine annotations are used for training.

\noindent $\bullet$
\textbf{ADE20K} is a dataset for the ILSVRC2016 Scene Parsing Challenge, which contains more than 20K annotated images of natural scenes. The annotations are provided for 150 semantic categories such as objects and object-parts. The dataset has training, validation, and testing sets with 20K/2K/3K images, respectively.

\noindent $\bullet$
\textbf{PASCAL-context} is a scene parsing dataset with 59 classes and one background class, which consists of 4,998 training images and 5,105 validation images. Following previous works, we evaluated our models on the 59 classes and excluded background class.

\vspace{0.1cm}
\noindent
\textbf{Road extraction:}
We evaluate our method on DeepGlobe dataset \cite{Demir2018}, which consists of satellite images and corresponding pixel-wise road mask annotations. The images has 50 cm spatial resolution and pixel size of $1,024\times 1,024$. Following \cite{Batra2019}, we split the dataset into training and validation with 4,696 and 1,530 images, where the performance is evaluated on road IoU and APLS metrics \cite{VanEtten2018APLS}. The APLS metric measures similarity between a predicted and a ground truth road graph.

\vspace{0.1cm}
\noindent
\textbf{Salient object detection:}
Following previous works \cite{Wu2019sod, Zhao2020sod}, we train our models on DUTS \cite{Wang2017sod} dataset, and evaluate the models on ECSSD \cite{Yan2013sod}, PASCAL-S \cite{Li2014sod}, DUT-OMRON \cite{Yang2013sod}, HKU-IS \cite{Li2015sod}, SOD \cite{Movahedi2010sod}, and DUTS \cite{Wang2017sod}.
For evaluation metrics, mean absolute error (MAE) and maximum F-measure (maxF) are used as in prior works.

\section{More results}
\label{sect:examples}
Table~\ref{tb:result_per_class} shows the per-class segmentation performance gain achieved by HG-Conv for Cityscapes dataset. We see that the HG-Conv performs well on both small and large objects.

Figures~\ref{fig:pred_seg}, \ref{fig:pred_road}, \ref{fig:vis_seg}, \ref{fig:vis_road}, \ref{fig:pred_sal}, and \ref{fig:vis_sal} show additional experimental results for the same problems discussed in the paper.
\tbHyperparam
\figArch
\figPredSeg
\figPredRoad
\figVisSeg
\figVisRoad
\figPredSal
\figVisSal